\def\year{2020}\relax
\documentclass[letterpaper]{article} 
\usepackage{aaai20}  
\usepackage{times}  
\usepackage{helvet} 
\usepackage{courier}  
\usepackage[hyphens]{url}  
\usepackage{graphicx} 
\usepackage{graphicx}  
\usepackage{tabularx}
\usepackage{latexsym}
\usepackage{url}
\usepackage{amsmath}
\usepackage{color}
\usepackage{epsfig}
\usepackage{algorithm}
\usepackage{algorithmic}
\usepackage{caption}

\usepackage{epstopdf}
\usepackage{epsfig}
\usepackage{amsmath}
\usepackage{enumitem}

\usepackage{adjustbox}
\usepackage{amssymb}
\usepackage{multirow}
\usepackage[flushleft]{threeparttable}
\usepackage{subfigure}

\DeclareMathAlphabet{\pazocal}{OMS}{zplm}{m}{n}

\newcommand{\figfref}[1]{Fig~\ref{#1}}
\newcommand{\figref}[1]{Fig.~\ref{#1}}
\newcommand{\tabref}[1]{Table.~\ref{#1}}

\newcommand{\eg}{{\it e.g. }}
\newcommand{\ie}{{\it i.e. }}

\newcommand{\citet}[1]{\citeauthor{#1} \shortcite{#1}}

\title{Hide-and-Tell: Learning to Bridge Photo Streams for Visual Storytelling}

\author{Yunjae Jung\textsuperscript{\rm 1}, Dahun Kim\textsuperscript{\rm 1}, Sanghyun Woo\textsuperscript{\rm 1}, Kyungsu Kim\textsuperscript{\rm 2}, Sungjin Kim\textsuperscript{\rm 2}, In So Kweon\textsuperscript{\rm 1}\\
\textsuperscript{\rm 1}Korea Advanced Institute of Science and Technology (KAIST),  Korea \\
\textsuperscript{\rm 2}Samsung Electronics Co., Ltd (Samsung Research), Korea
}
 
\begin{document}

\maketitle

\begin{abstract}
Visual storytelling is a task of creating a short story based on photo streams. Unlike existing visual captioning, storytelling aims to contain not only factual descriptions, but also human-like narration and semantics. However, the VIST dataset consists only of a small, fixed number of photos per story. Therefore, the main challenge of visual storytelling is to fill in the visual gap between photos with narrative and imaginative story. In this paper, we propose to explicitly learn to imagine a storyline that bridges the visual gap. During training, one or more photos is randomly omitted from the input stack, and we train the network to produce a full plausible story even with missing photo(s). Furthermore, we propose for visual storytelling a hide-and-tell model, which is designed to learn non-local relations across the photo streams and to refine and improve conventional RNN-based models. In experiments, we show that our scheme of \textit{hide-and-tell}, and the network design are indeed effective at storytelling, and that our model outperforms previous state-of-the-art methods in automatic metrics. Finally, we qualitatively show the learned ability to interpolate storyline over visual gaps.
\end{abstract}

\section{Introduction}
Recent deep learning based approaches have shown promising results for vision-to-language problems~\cite{vinyals2015show,karpathy2015deep,donahue2015long,yu2016video,pan2016hierarchical,gao2017video} that require the generation of text descriptions from given images or videos. Most existing methods have focused on giving direct and factual descriptions of visual content. While this is a promising first step, it is still challenging for artificial intelligence to connect vision with more naturalistic and human-like language. One emerging task proposed to take one step closer to human-level description is visual storytelling~\cite{huang2016visual}. Given a stream (set) of photos, this method aims to create a narrative, evaluative and imaginative story based on semantic visual understanding. While conventional visual descriptions are visually grounded, visual storytelling tries to describe contextual flow and overall situation across the photo stream, and so its output sentences can contain words for objects that do not even appear in the given image. Therefore, filling in the visual gap between the given photos with a subjective and imaginative story is the main challenge of visual storytelling.


\begin{figure}[t]
\begin{center}
\def\arraystretch{1.0}
\begin{tabular}{@{}c@{\hskip 0.005\textwidth }c@{\hskip 0.005\textwidth }c@{\hskip 0.005\textwidth }c@{\hskip 0.005\textwidth }c@{}}
\includegraphics[width=0.19\linewidth]{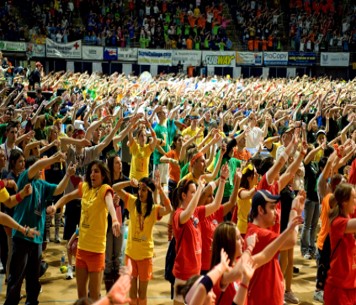} & \includegraphics[width=0.19\linewidth]{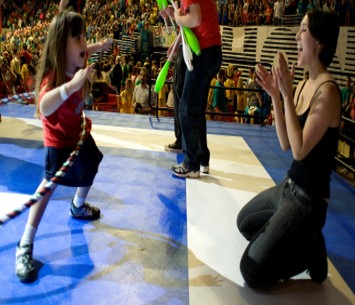} &
\includegraphics[width=0.19\linewidth]{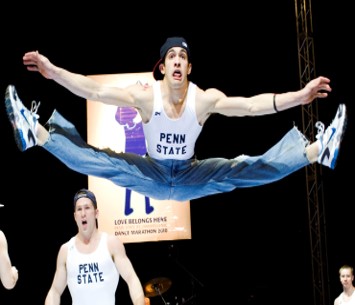} & \includegraphics[width=0.19\linewidth]{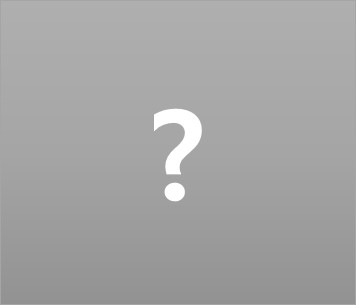} &
\includegraphics[width=0.19\linewidth]{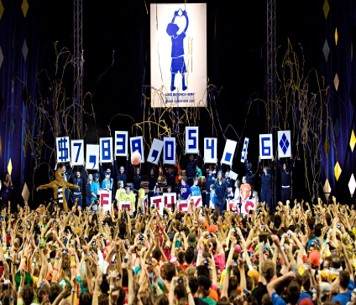} \\
(a) & (b) & (c) & (d) & (e) \\
\end{tabular}
\begin{tabular}{@{}c@{}}
\includegraphics[width=1.0\linewidth]{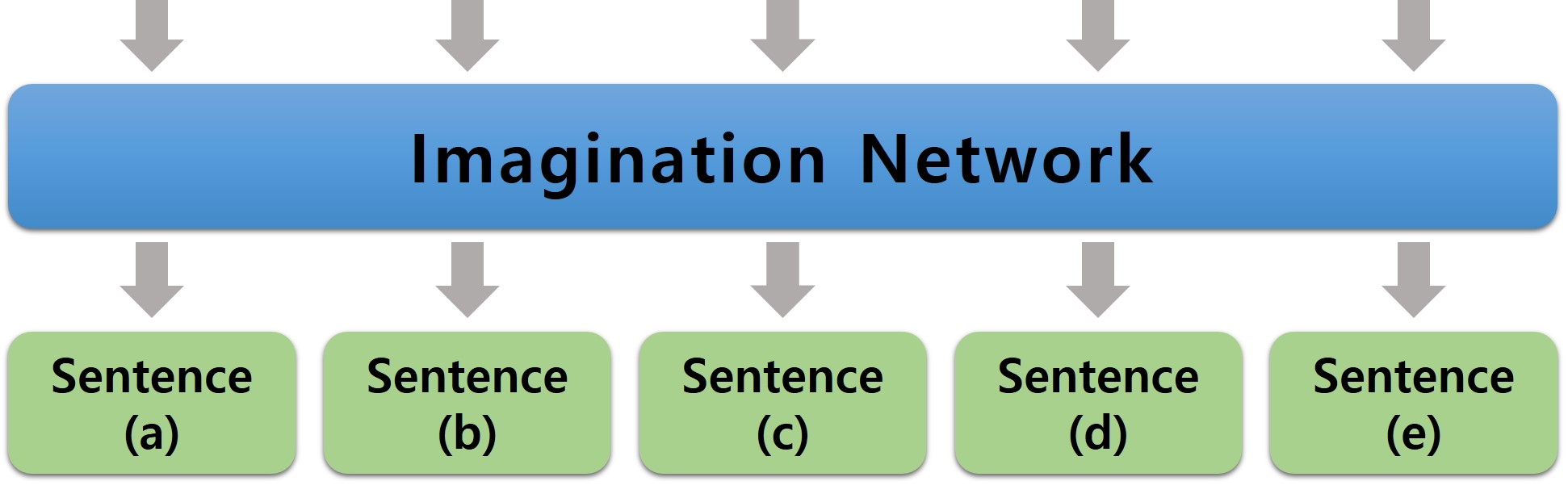} \\
\end{tabular}
\end{center}
\textbf{Generated Story:} (a) The fans were excited for the game. (b) There were many people there. (c) The lead singer performed a great performance. \textcolor{red}{(d) The game was very intense.} (e) This was a great game of the game.
\caption{\textbf{Example of our hide-and-tell prediction.} In this example, our hide-and-tell network takes four valid images and one black image. It is designed to learn contextual relations across the photo stream. Despite the hidden photo, our predicted sentence (d) \textit{"The game was very intense"} is semantically natural and plausible with the whole story context. }
\label{fig:teaser}
\end{figure}

In this paper, we propose to explicitly learn to imagine the storyline that bridges the visual gap. To this end, we present an auxiliary \textit{hide-and-tell} training task to learn such ability. As shown in \figref{fig:teaser}, one or more photos in the input stack are randomly masked during training. We train our model to produce a full, plausible story even with a missing photo(s). This image dropout in training encourages our model to describe what is happening \textit{in} the given stream of photos, as well as \textit{between} the photos. Since this story imagination task is an ill-posed problem, we follow curriculum learning, in which we start with an original setting in the early steps, and gradually increase the number of image dropout during training.

Furthermore, we propose an imagination network that is designed to learn non-local relations across photo streams to refine and improve, in a coarse-to-fine manner, the recurrent neural network (RNN) based baseline. We build upon a strong baseline model (XE-SS)~\cite{wang2018no} that has a CNN-RNN architecture and is trained with cross-entropy loss. Since we focus on learning contextual relations among all given photo slots, even those with missing photos, we propose to add a non-local (NL) layer~\cite{wang2018non} after the RNN block to refine long-range correlations across the photo streams. Our imagination network is designed with the first CNN block, and a stack of two RNN-NL blocks with a residual connection between; the following gated recurrent unit (GRU) outputs the final storyline.


In the experimental section, we evaluate our results with automatic metrics of BLEU~\cite{papineni2002bleu}, METEOR~\cite{banerjee2005meteor}, ROUGE~\cite{lin2004rouge}, and CIDEr~\cite{vedantam2015cider}. We conduct a quantitative ablation study to verify the contribution of each of the proposed design components. Also, we compare our imagination network with existing state-of-the-art models for visual storytelling. By conducting a user study, we show that our results are qualitatively better than the baselines. Another user study demonstrates that our hide-and-tell network is able to predict a plausible overall storyline even with missing photos. Finally, we introduce a new task of story interpolation, which involves predicting language descriptions not only for the given images, but also for gaps between the images.

Our contributions are summarized as follows.
\begin{itemize}
\item We propose a novel hide-and-tell training scheme that is effective for learning imaginative ability for the task of visual storytelling
\item We also propose an imagination network design that improves over the conventional RNN-based baseline.
\item Our proposed model achieves state-of-the-art visual storytelling performances in terms of automatic metrics.
\item We qualitatively show that our network faithfully completes the storyline even with corrupted input photo stream, and is able to predict inter-photo stories.
\end{itemize}

\section{Related Work}


\paragraph{Visual Storytelling}
\quad

\noindent
Visual storytelling is a problem of generating human-like descriptions with images selected from a photo album. Unlike conventional captioning tasks, visual storytelling aims to create a subjective and imaginative story with semantic understanding in the scenes. Early work~\cite{park2015expressing} exploits user annotation from blog posts. Newly released VIST~\cite{huang2016visual} dataset with a narrative story leads to several follow-up studies. Approaches with hierarchical concept~\cite{yu2017hierarchically,wang2019hierarchical} are proposed. And~\citet{wang2018show,wang2018no} formulate a visual storytelling task using adversarial reinforcement learning methods.

\begin{figure*}[t]
\centering
\includegraphics[width=1.0\textwidth]{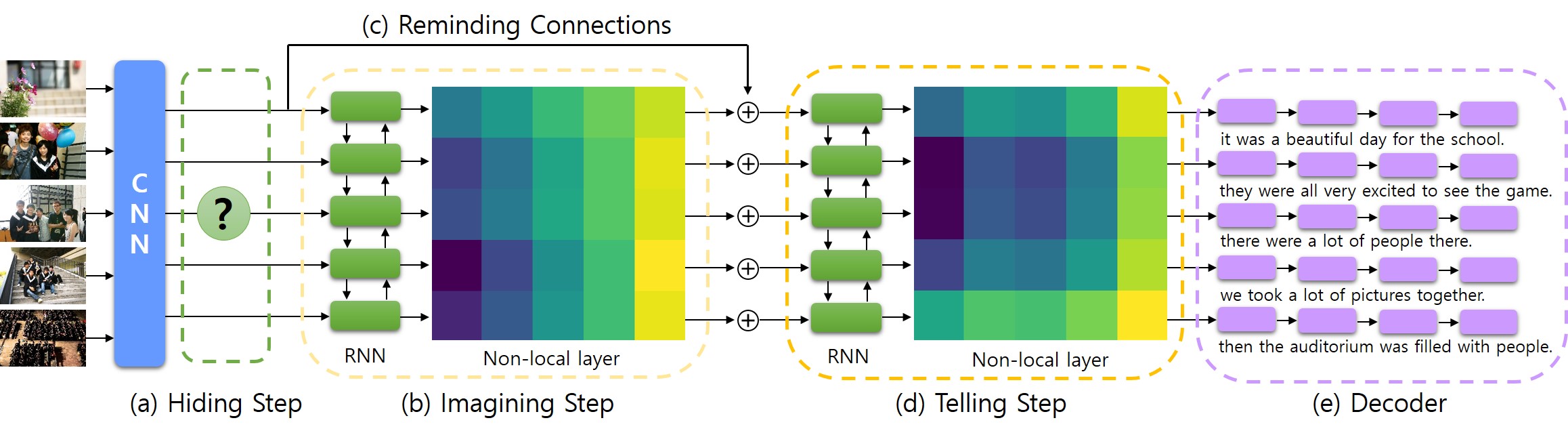}
\caption{\textbf{The overall architecture of our imagination network.} (a) In hiding step, one (or two) of the five inputs are randomly omitted by zero-masking. (b) In the imagining step, the inter-frame relations are roughly captured by the proposed imagination network which is composed of RNN (\eg GRU) and non-local self-attention module. (c) By utilizing residual connections, the imagination network can focus on recovering the blinded features. (d) The telling step refines the whole features using the same architecture with that of the imagining step, while the parameters are not shared between them. (e) The decoder generates a final story that describes the whole images.}
\label{fig:overview}
\end{figure*}

\paragraph{Overcoming Bias}
\quad

\noindent
Overfitting is a long-stand problem of deep neural network which causes difficulty in test cases. 
To alleviate this problem, dropout~\cite{srivastava2014dropout} is widely adopted. During training, it randomly drops weights in the neural networks to avoid severe co-adapting. For language models, a similar approach~\cite{ji2015blackout} named blackout is proposed to increase stability and efficiency. While dropout is often used at hidden layers of networks, blackout only targets output layers. Recently, for captioning models, \citet{burns2018women} tries to overcome bias in gender-specific words by occluding gender evidence in training images.

Illustrated hiding methods motivate our input blind learning scheme. It randomly obscures one or two images from the input in the training stage. 
Since the VIST dataset has a fixed number of input images as five, there can be overfitting in learning relations among images. From this point of view, our hide-and-tell concept gains performance improvement from the perspective of regularization.
Also, visual storytelling aims to generate subjective and imaginative descriptions unlike conventional captioning. In that regard, our approach has the advantage that the network learns to imagine the skipped input.

\paragraph{Curriculum Learning}
\quad

\noindent
Inspired by the human learning process, \citet{bengio2009curriculum} proposed curriculum learning which starts from relatively easy task and gradually increases the difficulty of training. It benefits both performance improvement and speed of convergence in various deep learning tasks such as optical flow~\cite{ilg2017flownet}, visual question answering~\cite{misra2018learning}, and image captioning~\cite{ren2017deep}. We also exploit curriculum learning by scheduling the difficulty of a task. At the early steps of training, there is no obscured input. Then, one of the five input images is omitted in the later step. Lastly, two of the five input images are hidden. If a validation loss is saturated, each step goes into the next step.

\paragraph{Relational Embedding}
\quad

\noindent
Recently, a non-local neural network~\cite{wang2018non} is proposed to capture long-range dependencies with self-attention. in other words, it computes the relations along with spatio-temporal spaces. 
Also, the non-local layer is a flexible network that can be well suited to both convolution layers and recurrent networks.
It is widely used to vision tasks such as scene graph generation~\cite{woo2018linknet}, image generation~\cite{zhang2018self}, 
and NLP tasks such as image and video captioning~\cite{gao2019hierarchical}, text classification and sequence labeling~\cite{liu2018contextualized}. 
We also exploit the self-attention mechanism of the non-local layer to our networks which try to imagine a story for the hidden images by learning relations between images.

\section{Proposed Approach}
An overview of the proposed imagination network is shown in \figref{fig:overview}. 
Given five input images \textit{$ I = \{I_{1}, I_{2}, I_{3}, I_{4}, I_{5}\} $}, the model outputs five corresponding sentences \textit{$ S = \{s_{1}, s_{2}, s_{3}, s_{4}, s_{5}\} $}.
Each sentence consists of several words \textit{$ W = \{w_{1}, w_{2}, \cdot\cdot\cdot , w_{T}\} $}, where $T$ denotes the length of the sentence.

Our model operates in three steps: \textbf{Hide}, \textbf{Imagine}, and \textbf{Tell}. 
After the first convolutional layer, which extracts visual features from each input photo, the hiding step randomly blinds one or two image features. It is implemented by setting the selected feature values to 0. During training, we employ a curriculum learning scheme, which starts with a normal setting (without hiding) and gradually increases the number of hidden images to two image features. (\ie 0 to 2). In our preliminary experiment, we found that blinding three or more image features does not provide further performance improvement.

Second, the imagining step consists of the aforementioned RNN-NL block. The goal of this step is to make a coarse initial prediction for the omitted features. Together with a residual connection from the CNN feature stack, this step captures contextual relations between the known image features, focusing on recovering the missing features. Finally, the telling step takes the feature stack from the previous imagining step, and refines the relational embedding to capture more concrete semantics throughout the photo stream. The RNN-NL block in this step shares the same architecture as that of imagining step, while the parameters are not shared. The refined feature stack is fed into the decoder to generate the final language output.


\subsection{Hide-and-Tell Learning}
\paragraph{Hiding Step}
\quad

\noindent
The input photo stream \textit{$ I = \{I_{1}, I_{2}, I_{3}, I_{4}, I_{5}\} $} is fed into the pre-trained CNN layer, which extracts high-level image features \textit{$ F = \{f_{1}, f_{2}, f_{3}, f_{4}, f_{5}\} $}. As shown in \figref{fig:overview}-(a), one or two of the features $F$ are randomly dropped in the hiding step. Although the missing information makes the reconstruction task an ill-posed problem, the method of hiding not only has a regularization effect but also helps our model to learn the contextual relations that lead to a performance gain in testing.

\begin{equation}
F_{blind} = \{ b_{1} f_{1}, b_{2} f_{2}, \cdot\cdot\cdot, b_{5} f_{5} \}, \quad b_{n} \in \{0, 1\}
\label{equ:blinding}
\end{equation}
where $n$ denotes the number of input images, $F_{blind}$ is a feature set including zero-masked features, $b_{n}$ is a masking weight which is randomly set during training.

\paragraph{Curriculum Learning}
\quad

\noindent
It is very challenging even for human intelligence to recover the missing features by using the neighboring photos in the same input stack. To ease the training difficulty in early steps, we adopt a curriculum learning scheme~\cite{bengio2009curriculum}. In early training, our imagination network is given fully visible photo stack (\ie $b_{total} = 0$). When the training loss becomes saturated, we start to hide one image feature from the input stack (\ie $b_{total} = 1$). Similarly, we proceed to hide two image features (\ie $ b_{total} = 2$) in the later steps.

\begin{eqnarray}
    b_{total} = 
\begin{cases}
0,& \text{if  } \text{epoch}  < \alpha \\
1,& \text{if  } \alpha \leq \text{epoch} < \beta \\
2,&  \text{otherwise}
\end{cases}
\label{equ:curriculum}
\end{eqnarray}

where $\alpha, \beta$ are hyper parameters which are empirically determined as the saturation point. The effect of curriculum learning is shown in the experiment section (\tabref{tab:curriculum}).

\begin{figure}[t]
\begin{center}
\def\arraystretch{1.0}
\begin{tabular}{@{}c@{\hskip 0.005\textwidth }c@{}}
\includegraphics[width=0.9\linewidth]{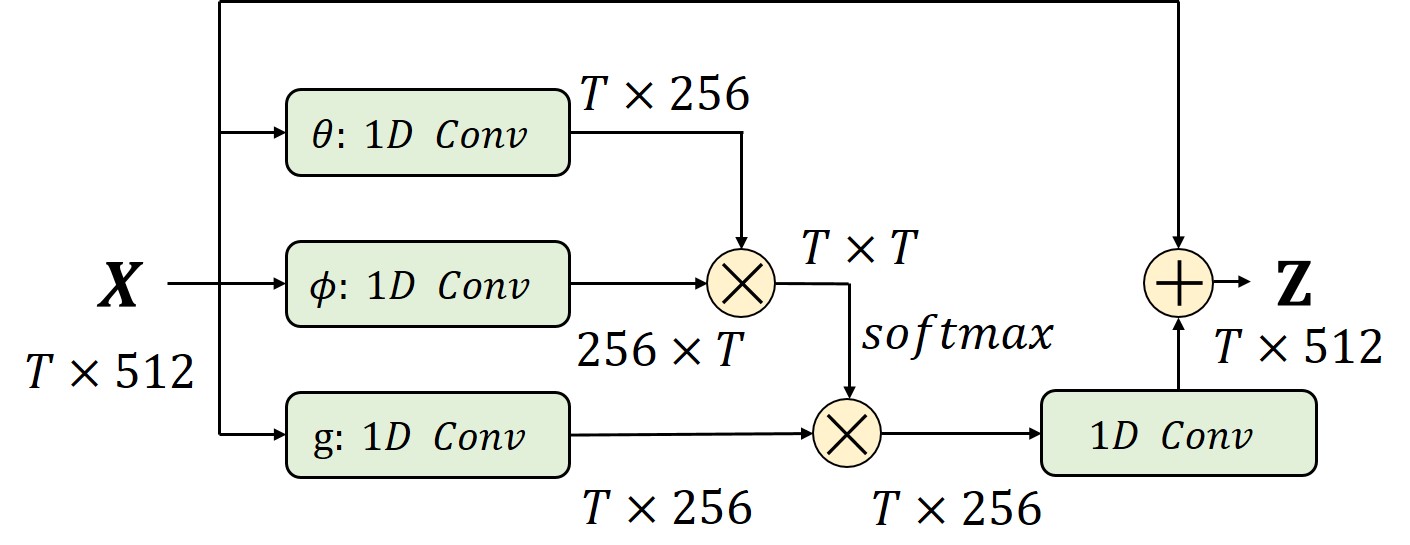} \\
\end{tabular}
\end{center}
\caption{\textbf{Relational embedding layer.} The features are first reshaped as a matrix form. Three parallel 1-D convolutions are used for feature embedding. The non-local operation starts with computing the correlation map ($T \times T$). It is produced by multiplying the output of $\theta$ and $\phi$ with the following $softmax$ normalization. The map is then multiplied with the output of $g$. The residual connection from $X$ to $Z$ allows the non-local block to be incorporated into existing RNN layers.}
\label{fig:non-local}
\end{figure}

\subsection{Imagination Network}
Our imagination network (INet) is designed to learn contextual relations between images in the input stack, and to generate human-like stories even with omitted photo(s). Following a coarse-to-fine pipeline, our network includes a \textit{coarse} imagining step and a \textit{fine} telling step that correspond to \figref{fig:overview}-(b) and (d), respectively. We use RNN-NL block in both steps.

\paragraph{Imagining Step}
\quad

\noindent
In the imagining step, $F_{blind}$ is fed into the bidirectional gated recurrent unit (GRU). In the forward direction, Bi-GRU takes $F_{blind} = \{ b_{1} f_{1}, b_{2} f_{2}, \cdot\cdot\cdot, b_{5} f_{5} \}$ and embeds according hidden states $( \overrightarrow{h_1}, \cdot\cdot\cdot, \overrightarrow{h_5} )$. Then, in the backward direction, reversed hidden states $( \overleftarrow{h_1}, \cdot\cdot\cdot, \overleftarrow{h_5} )$ are generated. The hidden states are concatenated into $h_i = \big[ \overrightarrow{h_i} ; \overleftarrow{h_i} \big]$.

To model non-local relations between the images, we employ an embedded Gaussian version of a non-local neural network~\cite{vaswani2017attention,wang2018non}. As illustrated in \figfref{fig:non-local}, our relational embedding is different from most existing non-local approaches in that it considers each input image feature as one element and focuses on the relations between the photo streams. Detailed equations for relational embedding are as follows:

\begin{eqnarray}
y = softmax(x^{T} \, W_{\theta}^{T} \, W_{\phi} \, x) \,  g(x), \label{equ:non-local1}\\
Z = W_{z} \, y + x,
\label{equ:non-local2}
\end{eqnarray}
where, $x$ is the hidden states from GRU (\ie $h_i$), and each $W$ denotes 1D convolution layers because our approach does not consider the spatial dimension of each input image, but considers each image feature as one element.

Inspired from residual shortcuts~\cite{he2016deep}, a reminding connection is added to connect initial CNN features to the end of the first RNN-NL block. By adding $F_{blind}$ to $Z$, which is the output of the relational embedding layer, the first RNN-NL block is encouraged to focus on recovering the missing features.

\begin{eqnarray}
F_{reminded} = F_{blind} + Z.
\label{equ:remind}
\end{eqnarray}

\paragraph{Telling Step}
\quad

\noindent
In the telling step in \figref{fig:overview}-(d), the features from the previous imagining step, $F_{reminded}$, are fed into the second RNN-NL block, which shares the same architecture as the first block, but does not share the weight parameters. The features that have been hidden during the hiding step are now roughly reconstructed in the feature stack $F_{reminded}$, and the second RNN-NL block refines these features to allow more concrete and associative understanding of all the photos in the input stream. Thus, to make better language predictions, the second block focuses more on refining the features of all photo elements.

The decoder (\figref{fig:overview}-(e)) consists of GRU and generates sentences $ S = \{s_{1}, s_{2}, s_{3}, s_{4}, s_{5}\} $ for each input photos. In order to generate each sentence $S$, word $ W = \{w_{1}, w_{2}, \cdot\cdot\cdot , w_{T}\} $ are recurrently predicted in one-hot vector $v_{t}$ as follows:

\begin{eqnarray}
w_{t} = \text{GRU}(w_{t-1},[f_{tell} ; v_{t-1}] ), \\
v_{t} = softmax(W_{w} \, w_{t}),
\label{equ:decoder}
\end{eqnarray}
where $W_{w}$ denotes fully connected (FC) layer and non-linearity (\eg hyperbolic tangent function).

\section{Experiments}

\subsection{Experimental Setup}
\paragraph{Datasets}
\quad

\noindent
Our experiments are conducted on the VIST dataset which provides 210,819 unique photos from 10,117 Flickr albums for visual storytelling tasks. Given five input images selected from an album, corresponding five sentences annotated by users are provided as ground truth. For the fair comparison, we follow the conventional experimental settings used in existing methods~\cite{yu2017hierarchically,wang2018no}. Specifically, three broken images are excluded in our experiments. Also, the same number of training, validation, and test sets are used: 4,098, 4,988, and 5,050.

\paragraph{Evaluation Metrics}
\quad

\noindent
In order to quantitatively measure our method for storytelling, automatic metrics such as BLEU, ROUGE-L, CIDEr, and METEOR are adopted. We employ the same evaluation code used in existing methods~\cite{yu2017hierarchically,wang2018no}. Two sets of human-subjective studies are performed for further comparison.

\paragraph{Implementation Details}
\quad

\noindent
We reproduced XE-ss~\cite{wang2018no} and set as our baseline network. However, our approach is completely different from their adversarial reinforcement learning method except for the baseline (\ie XE-ss). ResNet-152~\cite{he2016deep} is used for the pre-trained CNN layer in \figref{fig:overview}. We empirically choose hyper parameters for curriculum learning; $\alpha = 50, \beta = 80$. The learning rate starts with $4e-4$, and it decays by half when the training difficulty is changed (\ie $\text{epoch} = \alpha \; \text{or} \; \beta$). Adam optimizer is used. For non-linearity in the network, ReLU~\cite{nair2010rectified} is used for pre-trained CNN layers and SELU~\cite{klambauer2017self} is employed for the imagining step and the telling step. In decoding stage, beam search is utilized with $\text{beam size}=3$. For fair experiments, we removed randomness along with different experiments by fixing a random seed. In other words, our experimental results do not rely on multiple trials. 

\begin{table}
\centering
\resizebox{1.0\linewidth}{!}
{%
\begin{tabular}{ l | c  c  c  c  c  c  c}
\hline
Method & B-1 & B-2 & B-3 & B-4 & M & R & C    \\
\hline
\hline
INet - B        & 63.7 & 39.1 & 23.0 & 13.9 & 35.1 & 29.2 & 9.9  \\
INet - N        & 64.4 & 39.8 & 23.6 & 14.3 & 35.4 & 29.6 & 9.4  \\
INet - R        & 63.5 & 39.0 & 22.9 & 13.9 & 35.0 & 29.4 & 9.2  \\
\hline
INet         & \textbf{64.4} & \textbf{40.1} & \textbf{23.9} & \textbf{14.7} & \textbf{35.6} & \textbf{29.7} & \textbf{10.0}  \\
\hline
\end{tabular}
}
\caption{\textbf{Ablation Study.} We block important components of INet to empirically verify its contributions to the final performance. INet-B, INet-N, and INet-R denote INet without blinding, non-local layers and the second RNN-NL block respectively.}
\label{tab:ablation}
\end{table}

\begin{table}
\centering
\resizebox{1.0\linewidth}{!}{%
\begin{tabular}{ l | c  c  c  c  c  c  c}
\hline
Methods & B-1 & B-2 & B-3 & B-4 & M & R & C    \\
\hline
\hline
0           & 63.7 & 39.1 & 23.0 & 13.9 & 35.1 & 29.2 & 9.9  \\
1           & 64.2 & 39.8 & 23.7 & 14.6 & 35.5 & 29.7 & 9.9  \\
2           & 62.7 & 39.1 & 23.5 & 14.4 & 35.5 & 29.5 & 9.2  \\
(0, 1)      & 63.7 & 39.6 & 23.5 & 14.4 & 35.4 & \textbf{29.9} & 9.8  \\
(0, 1, 2)   & \textbf{64.4} & \textbf{40.1} & \textbf{23.9} & \textbf{14.7} & \textbf{35.6} & 29.7 & \textbf{10.0}  \\
\hline
\end{tabular}
}
\caption{\textbf{Curriculum learning.} To show the effect of the curriculum learning, we experiment by varying the number of image dropouts for each item. The left column denotes the number of hidden input features during training. The (0, 1, 2) means that the number of hiding increases from 0 to 2. And the 0 denotes the number of hiding is fixed to 0.}
\label{tab:curriculum}
\end{table}

\subsection{Quantitative Results}
\paragraph{Ablation Study}
\quad

\noindent
We conduct an ablation study to demonstrate the effects of different components of our method in \tabref{tab:ablation}.

Our model has three distinctive components; hiding step, non-local attention layer, and imagination network.
We investigate the importance of each component. 
If we provide non-blinded (\ie fully-visible) input features to the model, the model loses the regularization effects. 
We call this model as \textit{INet-B}.
If we omit the non-local attention layers, the network should only rely on the recurrent neural network (RNN) to capture the inter-frame relations, missing the complementary effects of the non-local relations. 
We named this model as \textit{INet-N}. 
If we do not use the telling step, the model only has one imagining step which shows insufficient performance to generate more concrete sentences on the photo stream. 
We named this model as \textit{INet-R}.

In all ablation setups, we observe performance drops. 
The model \textit{INet-B} shows that simply using all the image features is not enough to get good results as it is prone to overfitting. This shows the effectiveness of the proposed \textit{hide-and-tell} learning scheme. The model \textit{INet-N} suffers from its structural limitation as it purely depends on the recurrent neural networks for modeling the inter-frame relationship, and has difficulty handling complex relations between the frames. The result of model \textit{INet-R} implies that the refinement stage after the first imagination step is crucial.

\begin{table}
\centering
\resizebox{1.0\linewidth}{!}{%
\begin{tabular}{ l | c  c  c  c  c  c  c}
\hline
Method & B-1 & B-2 & B-3 & B-4 & M & R & C    \\
\hline
\hline
Huang et al. &  -   &  -   &   -  & -    & 31.4 &   -  &  -  \\
Yu et al.    &  -   &  -   & 21.0 & -    & 34.1 & 29.5 & 7.5  \\
HPSR         & 61.9 & 37.9 & 21.5 & 12.2 & 34.4 & \textbf{31.2} & 8.0  \\
GAN          & 62.8 & 38.8 & 23.0 & 14.0 & 35.0 & 29.5 & 9.0  \\
XE-ss        & 62.3 & 38.2 & 22.5 & 13.7 & 34.8 & 29.7 & 8.7  \\
AREL (best)  & 63.8 & 39.1 & 23.2 & 14.1 & 35.0 & 29.5 & 9.4  \\
HSRL         & -    & -    & -    & 12.3 & 35.2 & 30.8 & \textbf{10.7}  \\
\hline
INet         & \textbf{64.4} & \textbf{40.1} & \textbf{23.9} & \textbf{14.7} & \textbf{35.6} & 29.7 & 10.0 \\
\hline
\end{tabular}
}
\caption{\textbf{Comparison to Existing Methods.} Following automatic metrics are used: BLEU (B), METEOR (M), ROUGE-L (R), and CIDEr (C). 
The result shows that our approach achieves new state-of-the-art result. }
\label{tab:sota}
\end{table}

\begin{figure*}
\centering
\includegraphics[width=0.9\textwidth]{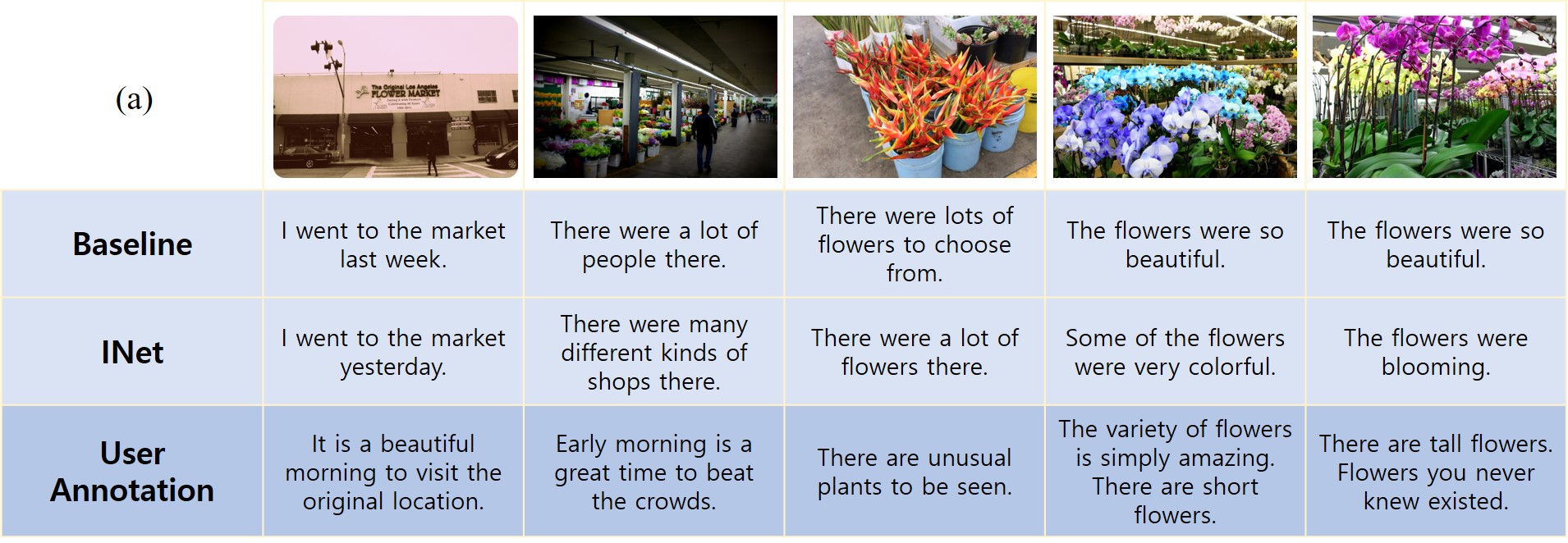} \\
\includegraphics[width=0.9\textwidth]{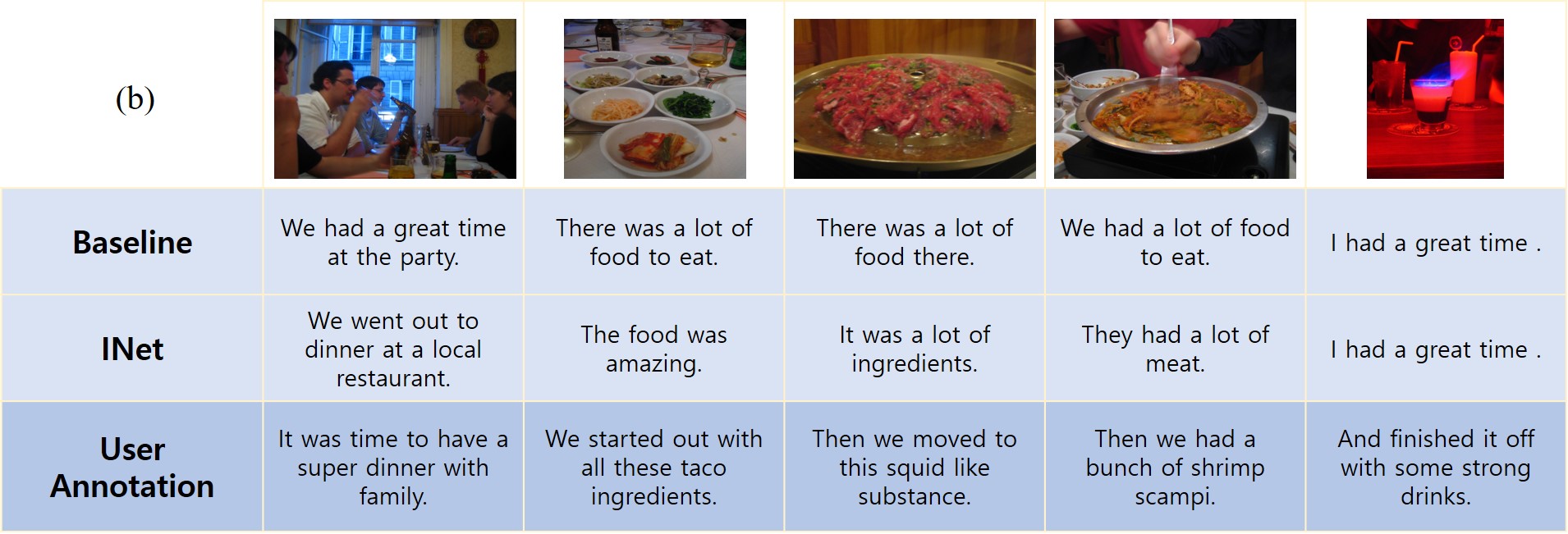} 
\caption{\textbf{Non-hiding test.} We qualitatively compare the results of baseline and the results of INet using all input images without hiding in the inference stage. (a) The upper example. (b) The lower example.  }
\label{fig:vs_base}
\centering
\includegraphics[width=0.9\textwidth]{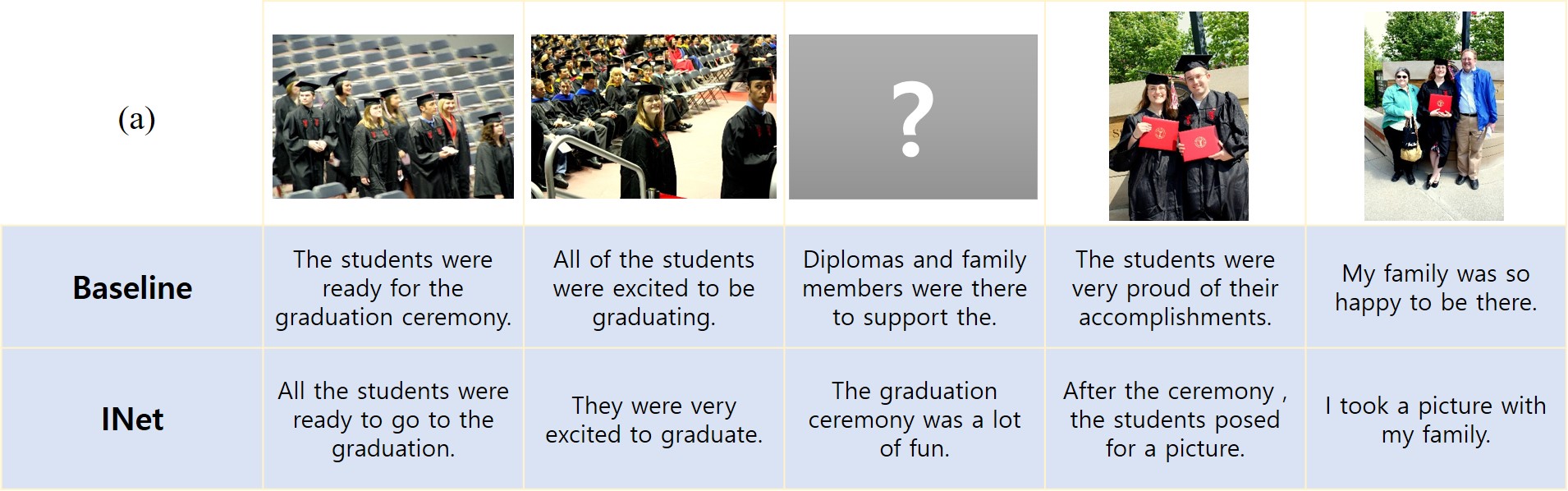} \\
\includegraphics[width=0.9\textwidth]{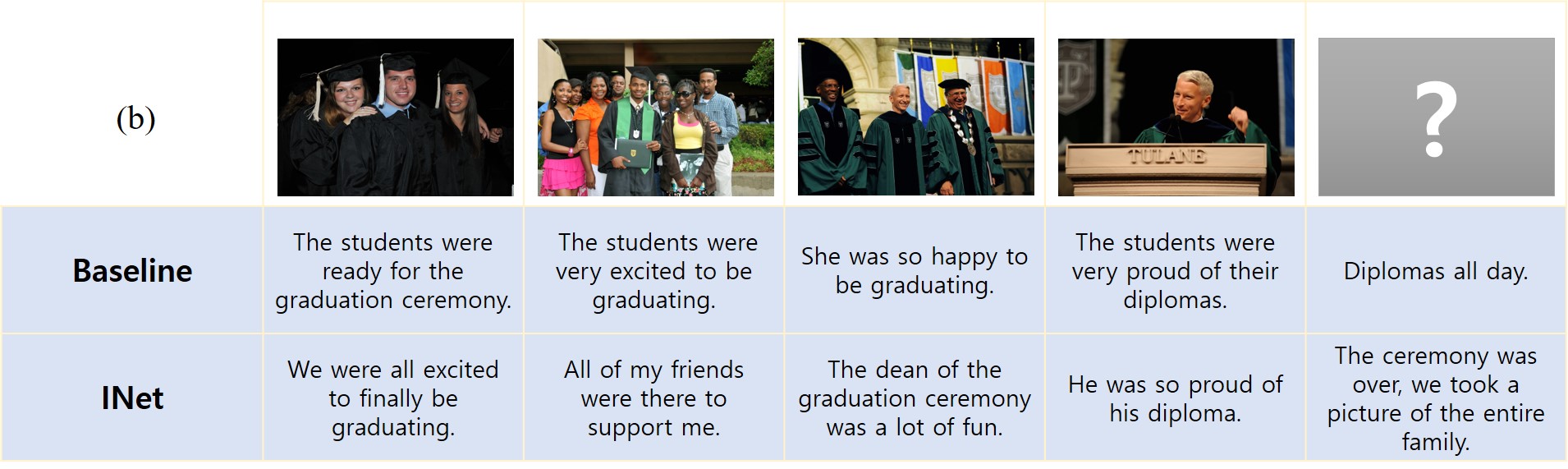} 
\caption{\textbf{Hiding test.} For the obscured input, we qualitatively show the results of the baseline and the results of INet. A story for the hidden image (\ie the third image in (a)) is also generated.
Unlike \figref{fig:vs_base}, user annotation is skipped in this experiment because users already know which input image is blinded. }
\label{fig:blind}
\end{figure*}

\paragraph{Comparison to Existing Methods}
\quad

\noindent
We compare our method with state-of-the-art methods~\cite{huang2016visual,yu2017hierarchically,wang2018no,wang2019hierarchical,huang2019hierarchically} in \tabref{tab:sota}. Our approach achieves the best results in BLEU and METEOR metrics. Compared with previous approaches, our approach could better handle complex sentences. However,
evaluation metrics are not perfect as there are many reasonable solutions for the narrative story generation. Therefore, we perform a user study and compare our approach with the strongest state-of-the-art baseline~\cite{wang2018no}. For each user study (\tabref{tab:user-1}, \tabref{tab:user-2}), thirty participants answered twenty five queries.
As shown in ~\tabref{tab:user-1}, we see that our approach significantly outperforms the baseline, implying that our method produces much more human-like narrations.

\begin{table}
\centering
\resizebox{0.7\linewidth}{!}{%
\begin{tabular}{c | c | c}
\hline
 XE-ss  &  Hide-and-tell  & Tie    \\
\hline
  24.7 \%       &  55.2 \%    &     20.1 \%   \\
\hline
\end{tabular}
}
\caption{\textbf{Baseline vs INet} without hiding in the test.  }
\label{tab:user-1}
\end{table}

\begin{table}
\centering
\resizebox{0.7\linewidth}{!}{%
\begin{tabular}{c | c | c}
\hline
 Full input &  Hidden input  &  Tie    \\
\hline
  30.9 \%       &  40.5 \%    &     28.6 \%   \\
\hline
\end{tabular}
}
\caption{\textbf{ INet vs $\text{INet}_{\text{hidden}}$.} In inference stage, we compare the story generated by INet with and without hidden images.}
\label{tab:user-2}
\end{table}

\begin{figure*}[t]
\centering
\includegraphics[width=0.75\textwidth]{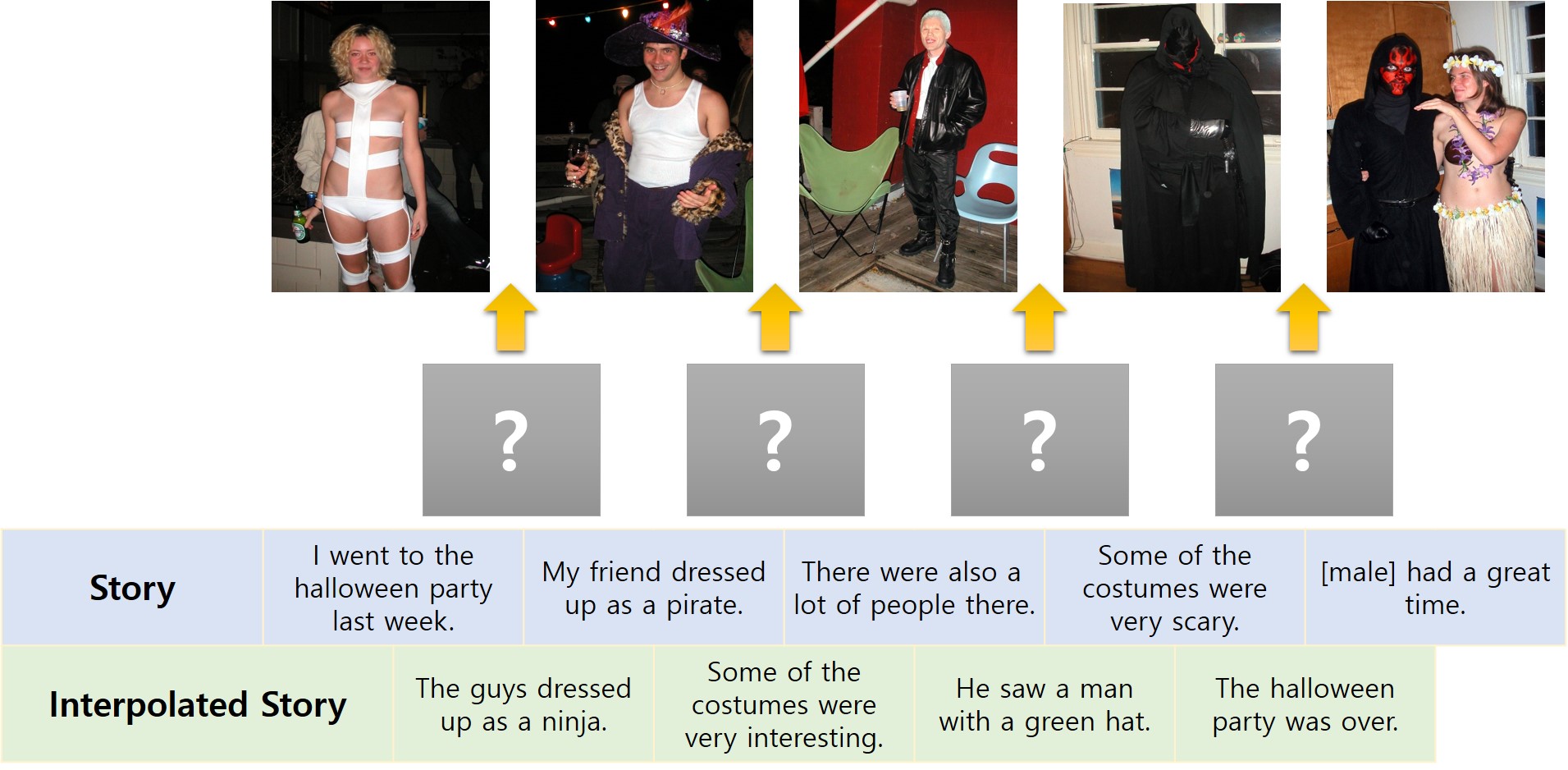}
\caption{\textbf{Story Interpolation.} Given five input images provided in the VIST dataset, we insert black images in between the five images. Our model is asked to predict the sentence descriptions for both valid and black images. The generated sentences are plausible, and the storyline shows natural contextual flow.}
\label{fig:interpolation}
\end{figure*}

\subsection{Qualitative Results}

\paragraph{Non-hiding Test}
\quad

\noindent
We qualitatively compare our model with the baseline~\cite{wang2018no} in \figref{fig:vs_base}. We can observe that our model produces more diverse and comprehensive expressions. 
For example, in \figref{fig:vs_base}-(a), the repeated sentences (\eg \textit{"The flowers were so beautiful"}) are generated by the baseline, whereas,  the results of ours show a wide variety of sentences (\eg \textit{"Some of the flowers were very colorful."}, \textit{"The flowers were blooming."}). Moreover, there exists an apparent gap in depicting the picture. For describing the second photo in \figref{fig:vs_base}-(a), ours \textit{"There were many different kinds of shops there"} is a better representation than baseline's \textit{"There were a lot of people there"}. We observe a similar phenomenon in the example (b) as well. While the baseline repeats the same expressions such as \textit{"There was a lot of food"}, our network generates a wide variety of descriptions such as \textit{"food", "ingredients", "meat"}.
The qualitative results above demonstrate again that our method greatly improves over the strongest baseline~\cite{wang2018no}.

\paragraph{Hiding Test}
\quad

\noindent
In this experiment, we explore the strength of INet by hiding the input images in testing.
As shown in \figref{fig:blind}, one of the five input images is omitted. 
Specifically, the third and fifth image are masked in \figref{fig:blind}-(a) and \figref{fig:blind}-(b) respectively.
We then show the story generated by ours and the baseline~\cite{wang2018no}.
We can clearly see that our method produces a much more natural story and well captures the associative relations between the images. For example, the results of the baseline do not even form a sentence such as (\eg \textit{"Diplomas and family members were there to support the."} or \textit{Diplomas all day.}). On the other hand, the results of ours not only well maintains the global coherency over the sentences and are more locally consistent with neighboring sentences (\eg \textit{"The graduation \textbf{ceremony} was a lot of fun."} and \textit{"\textbf{After the ceremony}, the students posed for a picture."}). 

In \tabref{tab:user-2}, we show that INet with one hidden image can generate a more human-like story than the INet without any hidden images. Thanks to the proposed hide-and-tell learning scheme, our INet is equipped with a strong imagination ability regardless of the input image masking.

\paragraph{Story Interpolation}
\quad

\noindent
The story interpolation is a newly proposed task in this paper. It aims to interpolate the story by predicting sentences in between the given photo stream. Since the photo stream has temporally sparse images, the current task of visual storytelling has limited expressiveness. 
However, the proposed story interpolation task can make the whole story more specific and concrete.

As illustrated in \figref{fig:interpolation}, a story for given five input images is generated. Additionally, the inter-story for inserted black images is also created with four sentences. The results of interpolation look thoroughly maintaining both global contexts over the whole situation and local smoothness with adjacent sentences. For instance, the generated sentence \textit{"The Halloween party was over."} maintains both the global context of whole situation (\ie \textit{halloween party}) and local smoothness (\ie \textit{party was over}) preceded by \textit{'[male] had a great time."}. 

Motivated by the importance of imagination in the visual storytelling task, we extend our blinding test (\figref{equ:blinding}) to the story interpolation task. While the blinding test recovers a story for the hidden input, story interpolation generates inter-story (\ie five plus four, total nine sentences).
Since creating a story by looking only at surrounding images without corresponding input obviously requires imagination, our hide-and-tell approach faithfully performs well due to our new learning scheme and network design.

\section{Conclusion}
In this paper, we propose the hide-and-tell learning scheme with imagination network for visual storytelling task which addresses subjective and imaginative descriptions. First, input hiding block omits an image from an input photo stream. Then, in imagining block, features of the hidden image are predicted by associating inter-photo relations with RNN and 1D convolution-based non-local layer. At the last, concrete relations between images are refined to generate sentences in the decoder. In experiments, our approach achieves state-of-the-art performance both in automatic metrics and human-subjective user studies. Finally, we propose a novel story interpolation task and show that our model well imagines the inter-story between given photo streams.

\clearpage

\bibliography{aaai-2020}
\bibliographystyle{aaai}

\end{document}